\documentclass{article}

\usepackage[final,nonatbib]{nips_2017}

\usepackage[utf8]{inputenc} 
\usepackage[T1]{fontenc}    
\usepackage{hyperref}       
\usepackage{url}            
\usepackage{booktabs}       
\usepackage{amsfonts}       
\usepackage{nicefrac}       
\usepackage{microtype}      
\usepackage{mathtools}
\usepackage{algorithm}
\usepackage{algpseudocode}
\usepackage{tikz}
\usepackage{pgfplots}
\usepackage{soul}
\usepackage{multirow}
\usepackage{amsthm}
\usepackage[square,sort,comma,numbers]{natbib}

\newtheorem{lemma}{Lemma}
\newtheorem{theorem}{Theorem}

\title{A Streaming Algorithm for Graph Clustering}

\author{
  Alexandre Hollocou\\
  INRIA\\
  Paris, France\\
  \texttt{alexandre.hollocou@inria.fr} \\
  \And
  Julien Maudet \\
  Ecole Polytechnique\\
  Palaiseau, France \\
  \texttt{julien.maudet@polytechnique.edu} \\
  \And
  Thomas Bonald \\
  Telecom-Paristech\\
  Paris, France \\
  \texttt{thomas.bonald@telecom-paristech.fr} \\
  \And
  Marc Lelarge \\
  INRIA-ENS\\
  Paris, France \\
  \texttt{marc.lelarge@ens.fr} \\
}

\begin{document}

\maketitle

\begin{abstract}
We introduce a novel algorithm
to perform graph clustering in the edge streaming setting.
In this model, the graph is presented as a sequence of edges
that can be processed strictly once.
Our streaming algorithm
has an extremely low memory
footprint as it stores only three integers per node
and does not keep any edge in memory.
We provide a theoretical justification of the design of the algorithm
based on the \emph{modularity} function, which is a usual metric
to evaluate the quality of a graph partition.
We perform experiments on massive real-life
graphs ranging from one million to more than one billion edges and we show that
this new algorithm runs more than ten times faster than existing algorithms
and leads to similar or better detection scores on the largest graphs.
\end{abstract}

\section{Introduction}\label{introduction}

\label{section-intro}

\subsection{Motivations}\label{motivations}

Graphs arise in a wide range of fields from biology
\citep{palla2005uncovering} to social media
\citep{mislove2007measurement} or web analysis
\citep{flake2000efficient}\citep{pastor2007evolution}. In most of these
graphs, we observe groups of nodes that are densely connected between
each other and sparsely connected to the rest of the graph. One of the
most fundamental problems in the study of such graphs consists in
identifying these dense clusters of nodes. This problem is commonly
referred to as \emph{graph clustering} or \emph{community detection}.

A major challenge for community detection algorithms is their ability to
process very large graphs that are commonly observed in numerous
fields. For instance, social networks have typically millions of nodes
and billions of edges (e.g. Friendster \citep{mislove2007measurement}).
Many algorithms have been proposed during the last ten years, using
various techniques ranging from combinatorial optimization to spectral
analysis \citep{lancichinetti2009community}. Most of them fail to scale
to such large real-life graphs \citep{prat2014high}
{and require the whole graph to be stored in memory, which
often represents a heavy constraint in practice.
Streaming the edges is a natural way to handle such massive graphs.
In this setting, the entire graph is not stored but processed edge by edge} \citep{mcgregor2014graph}.
{Note that the streaming approach is particularly relevant in most real-life applications
where graphs are fundamentally dynamic and edges naturally arrive in a streaming fashion.}

\subsection{Contributions}\label{contributions}

In this paper, we introduce a novel approach based on \emph{edge streams}
to detect communities in graphs.
The algorithm processes each edge strictly once.
When the graph is a multi-graph, in the sense that two nodes may be connected by more than one edge,
these edges are streamed independently.
The algorithm only stores three
integers for each node: its current community index, its current degree
(i.e.\ the number of adjacent edges that have already been processed), and the current community volume
(i.e.\ the sum of the degrees of all nodes in the community).
Hence, the time
complexity of the algorithm is linear in the number of edges and its
space complexity is linear in the number of nodes. In the experimental
evaluation of the algorithm, we show that this streaming algorithm is able to
handle massive graphs \citep{yang2015defining} with low execution time
and memory consumption.

The algorithm takes only one integer parameter $v_{\mathrm{max}}$
and, for each arriving edge $(i,j)$ of the stream,
it uses a simple decision strategy based on this parameter and
the volumes of the communities of nodes $i$ and $j$.
We provide a theoretical analysis that justifies the form of this decision strategy
using the so-called \emph{modularity} of the clustering.
Modularity, that has been introduced by the physics community \citep{newman2006modularity},
is one of the most widely used quality function for graph clustering.
It measures the quality of a given partition based on the
comparison between the number of edges that are observed in each cluster
and the number of edges that would be observed if the edges were
randomly distributed.
In our analysis,
we show that, under certain assumptions,
the processing of each new edge by our algorithm leads to an increase in modularity.

\subsection{Related work}\label{related-work}

A number of algorithms have been developed for detecting communities in graphs
\citep{fortunato2010community}. Many rely on the
optimization of some objective function that measures the quality of the
detected communities. {Modularity and other metrics}, like
conductance, out-degree fraction and the clustering coefficient
\citep{yang2015defining} have been used with success. Other popular methods include
spectral clustering \citep{spielman2007spectral}\citep{von2007tutorial},
clique percolation \citep{palla2005uncovering}, statistical inference
\citep{lancichinetti2010statistical},
random walks \citep{pons2005computing}\citep{whang2013overlapping}
and matrix factorization \citep{yang2013overlapping}.
These techniques have
proved to be efficient but are often time-consuming and fail to scale to
large graphs \citep{prat2014high}.

The streaming approach has drawn considerable interest in network
analysis over the last decade. Within the data stream model, massive
graphs with potentially billions of edges can be processed without being
stored in memory \citep{mcgregor2014graph}. A lot of algorithm have been
proposed for different problems that arise in large graphs, such as
counting subgraphs \citep{bar2002reductions}\citep{buriol2006counting},
computing matchings
\citep{goel2012communication}\citep{feigenbaum2005graph}, finding the
minimum spanning tree \citep{elkin2006efficient}\citep{tarjan1983data}
or graph sparsification \citep{benczur1996approximating}. Different
types of data streams can be considered: \emph{insert-only streams},
where the stream is the unordered sequence of the network edges, or
\emph{dynamic graph streams}, where edges can both be added or deleted.
Many streaming algorithms rely on \emph{graph sketches} which store the
input in a memory-efficient way and are updated at each step
\citep{ahn2012graph}.

In this paper, we use the streaming setting to define a novel community
detection algorithm. We use an insert-only edge streams and
define a minimal sketch, by storing only three integers per node.

\subsection{Paper outline}\label{paper-outline}

The rest of the paper is organized as follows. We first describe our
streaming algorithm in Section \ref{section-algorithm}.
A theoretical analysis {of this algorithm} is
presented in Section \ref{section-analysis}.
{In Section} \ref{section-experiments}, we evaluate experimentally the
performance of {our approach} on real-life graphs and compare it to
state-of-the-art algorithms.
Section \ref{section-conclusion} concludes the
paper.

\section{A streaming algorithm for community
detection}\label{a-streaming-algorithm-for-community-detection}

\label{section-algorithm}

In this section, we define {a novel} streaming algorithm for community
detection in graphs.

\subsection{Streaming setting}\label{notations}

We are given an undirected and unweighted multi-graph $G(V,E)$ where $V$
is the set of vertices and $E$ is a multi-set of edges
(i.e. an edge $(i,j)$ can appear multiple times in $E$). We use $n$ to
denote the number of nodes and $m$ the number of edges.
We use $w_{ij}$ to denote the number of edges between $i$ and $j$
(and we set $w_{ij}=0$ if $(i,j)\notin E$).
We assume that there is no self-loop, that is $w_{ii} = 0$ for all $i\in V$.
We use $w_i$ to denote the degree $\sum_j w_{ij}$ of node $i$,
and $w=\sum_i w_i = 2m$ to denote the weight of the graph, corresponding to the total degree.
Given a set of nodes $C$, we use $\mathrm{Vol}(C)=\sum_{i\in C} w_i$ to denote
the volume of $C$.

We consider the following streaming framework:
we are given a stream $S=(e_1,\ldots,e_m)$, which is an order sequence
of the multi-set $E$.
Note that each edge $e=(i,j) \in E$ appears exactly $w_{ij}$ times in $S$.

\subsection{Intuition}\label{motivation}

Although there is no universal definition of what a community is, most
existing algorithms rely on the principle that nodes tend to be more
connected within a community than across communities. Hence, if we pick
uniformly at random an edge \(e\) in \(E\), this edge is more likely to
link nodes of the same community (i.e., \(e\) is an
\emph{intra-community} edge), than nodes from distinct communities
(i.e., \(e\) is an \emph{inter-community} edge).
Equivalently, {if we assume that edges arrive in a random order}, we expect many
intra-community edges to arrive before the inter-community edges.

{This observation is used to design a streaming algorithm.}
For each arriving edge \((i, j)\), the
algorithm places \(i\) and \(j\) in the same community if the edge
arrives \emph{early} (intra-community edge) and splits the nodes in
distinct communities otherwise (inter-community edge). In this
formulation, the notion of an \emph{early} edge is of course critical.
In the proposed algorithm, we consider that an edge \((i, j)\) arrives
\emph{early} if the current {volumes of the communities} of nodes \(i\) and \(j\), accounting
for previously arrived edges only, is low.

{More formally, the algorithm considers successively each edge of the stream
$S=(e_1,e_2,\ldots,e_m)$.} Each node is initially in its own
community. At time $t$, a new edge \(e_t = (i, j)\) arrives and the algorithm performs
one of the following actions:
(a) \(i\) joins the community of \(j\);
(b) \(j\) joins the community of \(i\);
(c) no action.

{The choice of the action depends on the \emph{updated}
community volumes $\mathrm{Vol}(C(i))$ and $\mathrm{Vol}(C(j))$
of the communities of $i$ and $j$, $C(i)$ and $C(j)$, i.e., the volumes computed using the
edges $e_{1},...,e_{t}$. If $\mathrm{Vol}(C(i))$ or $\mathrm{Vol}(C(j))$ is greater than a
given threshold $v_{\mathrm{max}}$, then we do nothing;
otherwise, the node belonging to the smallest community (in volume)
joins the community of the other node and the volumes are updated.}

\subsection{Algorithm}\label{algorithm}

{We define our streaming algorithm} in Algorithm \ref{scoda}. It takes the
list of edges of the graph and {one integer parameter $v_{\mathrm{max}} \ge 1$}.
The algorithm uses three dictionaries $d$, $c$ and $v$, initialized
with default value $0$.
At the end of the algorithm, {$d_i$ is the degree of node $i$,
$c_i$ the community of node $i$,
and $v_k$ is the volume of community $k$}.
When an edge with an unknown node arrives, let say $i$,
we give this node a new community index, $c_i \leftarrow k$,
and increment the index variable $k$ (which is initialized with $1$).
For each new edge $e=(i, j)$,
the degrees of \(i\) and \(j\) {and the volumes of communities $c_i$ and $c_j$ are updated}.
Then, if these {volumes} are
both lower than the threshold parameter {$v_{\mathrm{max}}$},
the node {in the community with the lowest volume} joins the community of the other node. Otherwise, the communities
remain unchanged.

\begin{algorithm}
\caption{{Streaming algorithm for clustering graph nodes}}
\label{scoda}
\begin{algorithmic}[1]
\Require
Stream of edges $S$ and
parameter $v_{\mathrm{max}} \ge 1$
\State $d,v,c \leftarrow$ dictionaries initialized with default value $0$
\State $k \leftarrow 1$ (\emph{new community index})
\For{$(i,j) \in S$}
    \If{$c_i=0$}
        $c_i \leftarrow k$ and $k\leftarrow k+1$
    \EndIf
    \If{$c_j=0$}
        $c_j \leftarrow k$ and $k\leftarrow k+1$
    \EndIf
    \State
    $d_i\leftarrow d_i + 1$ and $d_j \leftarrow d_j + 1$ (\emph{update degrees})
    \State
    $v_{c_i} \leftarrow v_{c_i} + 1$ and $v_{c_j} \leftarrow v_{c_j} + 1$ (\emph{update community volumes})
    \If{$v_{c_i} \leq v_{\mathrm{max}}$ and $v_{c_j} \leq v_{\mathrm{max}}$}
       \If{$v_{c_i} \leq v_{c_j}$} (\emph{$i$ joins the community of $j$})
       \State $v_{c_j} \leftarrow v_{c_j} + d_i$
       \State $v_{c_i} \leftarrow v_{c_i} - d_i$
       \State $c_{i} \leftarrow c_{j}$
       \Else\ (\emph{$j$ joins the community of $i$})
       \State $v_{c_i} \leftarrow v_{c_i} + d_j$
       \State $v_{c_j} \leftarrow v_{c_j} - d_j$
       \State $c_{j} \leftarrow c_{i}$
       \EndIf
    \EndIf
\EndFor
\State \Return $(c_i)_{i\in V}$
\end{algorithmic}
\end{algorithm}

{Observe that, in case of equality $v_{c_i}=v_{c_j}\le v_{\mathrm{max}}$,
$j$ joins the community of $i$.
Of course, this choice is arbitrary and can be made random
(e.g., $i$ joins the community of $j$ with probability
$1/2$ and $j$ joins the community of $i$ with probability
$1/2$).}

\subsection{Complexity}\label{complexity}

The main loop is linear in the number of edges in the stream.
Thus, the time complexity of the algorithm is linear in $m$.

Concerning the space complexity, we only use three dictionaries of integers $d$, $c$ and $v$,
of size $n$. Hence, the
space complexity of the algorithm is
\(3n \cdot \texttt{sizeOf(int)} = O(n)\).
Note that the algorithm does not need to
store the list of edges in memory,
which is the main benefit of the streaming approach.
To implement dictionaries with default value $0$ in practice,
we can use classic dictionaries or maps, and,
when an unknown key is requested, set the value relative to this key to $0$.
Note that, in Python, the \texttt{defaultdict} structure already allows us to exactly
implement dictionaries with 0 as a default value.

\subsection{Parameter setting}\label{degree-threshold}

{Note that the algorithm can be run once
with multiple values of parameter $v_{\mathrm{max}}$.
In this case, only arrays $c$ and $v$ need to be duplicated for each value of
$v_{\mathrm{max}}$.
{In this multi-parameter setting, we obtain multiple results $(c^a)_{1\leq a\leq A}$
at the end of the algorithm,
where $A$ is the number of distinct
values for the parameter $v_{\mathrm{max}}$.
Then, the best $c^a$ can be selected
by computing quality metrics that only use dictionaries $c^a$ and $v^a$.
In particular, we do not want to use metrics that requires the knowledge of the input graph.
For instance, common metrics \citep{yang2015defining} like entropy $H(v) = - \sum_k \frac{v_k}{w} \log\left( \frac{v_k}{w} \right)$
or average density $D(c,v) = \sum_{k: C_k \neq \emptyset} \frac{1}{|P|} \frac{v_k}{|C_k|(|C_k|-1)}$,
where $C_k$ is the set of nodes in community $k$ and $P$ is the set of all non-empty communities,
can be easily computed from each pair of dictionaries $(v^a,c^a)$, and be used to select the best result $c^a$
for $a=1,\ldots,A$.}
Note that modularity cannot be used here as its computation requires the knowledge of the whole graph.

\section{Theoretical analysis}\label{theoretical-analysis}
\label{section-analysis}

{In this section, we analyze the modularity optimization problem in the edge-streaming setting
and qualitatively justify the conditions on community volumes used in Algorithm \ref{scoda}.}
See Appendices A, B and C for complete proofs. .

\subsection{Modularity optimization in the streaming setting}

{Modularity is a quality metric that is widely used in graph clustering} \cite{newman2006modularity}.
{Given a partition $P$ of the nodes,
modularity is defined as}
\begin{equation*}
Q = \frac{1}{w} \sum_{i\in V} \sum_{j\in V} \left(w_{ij} - \frac{w_i w_j}{w}\right) \delta(i,j)
\end{equation*}
{where $\delta(i,j)=1$ if $i$ and $j$ belongs to the same community $C\in P$ and $0$ otherwise.
Modularity can be seen as the difference between two probabilities,}
\begin{center}
$Q=\mathbb{P}[(i,j)\sim E:C(i)=C(j)] - \mathbb{P}[(i,j)\sim \mathcal{N}: C(i)=C(j)]$,
\end{center}
{where $\mathbb{P}[(i,j)\sim E: C(i)=C(j)]$ corresponds to the probability
to choose an edge of $G$ uniformly at random between two nodes of the same community,
and $\mathbb{P}[(i,j)\sim \mathcal{N}: C(i)=C(j)]$ to the probability
to choose an edge between two nodes from the same community
in the so-called \emph{null model} $\mathcal{N}$, where an edge $(i,j)$
is chosen with a probability proportional to the degrees of $i$ and $j$.
A classic approach to cluster graphs consists in finding a partition that maximizes $Q$.
Many algorithms have been proposed to perform this task \citep{blondel2008fast,newman2004fast} but,
to the best of our knowledge, none can be applied to our streaming setting.

The modularity can be rewritten as}
\begin{equation*}
Q = \frac{1}{w} \left[
\sum_{i\in V} \sum_{j \in V} w_{ij} \delta(i,j)
- \sum_{C\in P} \frac{\mathrm{Vol}(C)^2}{w} \right].
\end{equation*}

In our streaming setting, we are given a stream $S=(e_1,\ldots,e_{m})$ of edges such that
edge $(i,j)$ appears $w_{ij}$ times in $S$.
We consider the situation where $t$ edges $e_1,\ldots,e_t$ from the stream $S=(e_1,\ldots,e_m)$
have already arrived, and where we have computed a partition $P_t=(C_1,\ldots,C_K)$ of the graph.
We define $S_t$ as $S_t = \{e_1,\ldots,e_t\}$,
and $Q_t$ as
\begin{equation*}
    Q_t =
\sum_{C\in P_t} \left[ 2 \mathrm{Int}_t(C) - \frac{(\mathrm{Vol}_t(C))^2}{w}\right]
\end{equation*}
where
$\mathrm{Int}_t(C) = \sum_{(i,j)\in S_t} 1_{i\in C} 1_{i\in C}$ and
$\mathrm{Vol}_t(C) = \sum_{(i,j) \in S_t} (1_{i\in C} + 1_{j\in C})$.
Note that there is no normalization factor $1/w$ in the definition of $Q_t$
as it has no impact on the optimization problem.

We do not store the edges of $S_t$ but we assume that we have kept updated
node degrees $w_t(i)=\sum_{(i',j') \in S_t} (1_{i'=i} + 1_{j'=i})$
and community volumes $\mathrm{Vol}_t(C_k)$ in a streaming fashion
as shown in Algorithm \ref{scoda}.
We consider the situation where a new edge $e_{t+1}=(i,j)$ arrives.
We want to make a decision that maximizes $Q_{t+1}$.

\subsection{Streaming decision}

We can express $Q_{t+1}$ in function of $Q_t$ as stated in Lemma \ref{lemma-Qt}.
\begin{lemma}
\label{lemma-Qt}
If $e_{t+1}=(i,j)$ and if $P_{t+1}=P_t$,
$Q_{t+1}$ can be expressed in function of $Q_t$ as follows
\begin{equation*}
Q_{t+1}
= Q_t + 2 \left[ \delta(i,j) - \frac{\mathrm{Vol}_t(C(i)) + \mathrm{Vol}_t(C(j)) + 1 + \delta(i,j)}{w} \right]
\end{equation*}
where $C(v)$ denotes the community of $v$ in $P_t$,
and $\delta(i,j)=1$ if $i$ and $j$ belongs to the same community of $P_t$ and $0$ otherwise.
\end{lemma}

We want to update the community membership of $i$ or $j$ with one of the following actions:
(a) $i$ joins the community of $j$;
(b) $j$ joins the community of $i$;
(c) $i$ and $j$ stays in their respective communities.
We consider the case where nodes $i$ and $j$ belongs to distinct communities of $P_t$,
since all three actions are identical if $i$ and $j$ belong to the same community.
We want to choose the action that maximizes $Q_{t+1}$,
but we have a typical streaming problem where we cannot evaluate the impact of action (a) or (b)
on $Q_t$ but only on the term that comes from the new edge $e_{t+1}$.

{Let us consider action (a), where $i$ joins the community of $j$.
We can assume that $\mathrm{Vol}_t(C(i)) \leq \mathrm{Vol}_t(C(j))$
without loss of generality (otherwise we can swap $i$ and $j$).
We are interested in $\Delta Q_{t+1} = Q_{t+1}^{(a)} - Q_{t+1}^{(c)}$,
which is the variation of $Q_{t+1}$
between the state where $i$ and $j$
are in their own communities and the state where $i$ has joined $C(j)$.
We have
$\Delta Q_{t+1} = \Delta Q_t +
2 [1 - (\mathrm{Vol}_t(C(j)) - \mathrm{Vol}_t(C(i)) + 2 w_i(t) + 1)/w]$
where $\Delta Q_t$ is the variation of $Q_t$.
Lemma} \ref{lemma-Delta-Qt}
{gives us an expression for this variation.}

\begin{lemma}
\label{lemma-Delta-Qt}
\begin{equation*}
\Delta Q_t = Q_t^{(a)} - Q_t^{(c)}
= 2 \left[
L_t(i,C(j)) -
L_t(i,C(i))
- \frac{(w_t(i))^2}{w} \right]
\end{equation*}
where
\begin{equation*}
L_t(i,C)
= \sum_{(i',j')\in S_t}
\left[ 1_{i'\in C} \left( 1_{j'=i} - \frac{w_t(i)}{w} \right)
+ 1_{j'\in C} \left( 1_{i'=i} - \frac{w_t(i)}{w} \right)
\right].
\end{equation*}
\end{lemma}

{We define $l_t(i,C)$ as
$l_t(i,C) = L_t(i,C)/ \mathrm{Vol}_t(C)$.
Then, we can easily show that $l_t(i,C) \in [-1,1]$
and $\mathbb{E}[l_t(i,C)] = 0$ if edges
of $S_t$ follow the null model presented above.
$L_t(i,C)$ measures the difference between the number of edges connecting node $i$ to community $C$
in the edge stream $S_t$, and the number of edges that we would observe in the null model.
It can be interpreted as a degree of attachment of node $i$ to community $C$.
Thus, $l_t(i,C)$ can be seen as a \emph{normalized degree of attachment of node $i$
to community $C$} in the edge stream $S_t$.
}

{Lemma} \ref{lemma-Qt} {and} \ref{lemma-Delta-Qt} {gives us
a sufficient condition presented in Theorem} \ref{theorem-increase-condition}
{in order to have a positive variation $\Delta Q_{t+1}$ of the modularity when $i$ joins $C(j)$.}
\begin{theorem}
\label{theorem-increase-condition}
If $\mathrm{Vol}_t(C(i)) \leq \mathrm{Vol}_t(C(j))$,
then:
\begin{equation*}
\mathrm{Vol}_t(C(j)) \leq v_t(i,j)
\quad \Longrightarrow \quad \Delta Q_{t+1} \geq 0
\end{equation*}
where
\begin{equation*}
v_t(i,j) =
\begin{cases}
\frac{1 - (w_t(i) + 1)^2/w}{l_t(i,C(i)) - l_t(i,C(j))} & \text{if $l_t(i,C(i)) \neq l_t(i,C(j))$}\\
+\infty & \text{otherwise.}
\end{cases}
\end{equation*}
\end{theorem}

{Thus, we see that, if we have
$v_{\mathrm{max}} \leq v_t(i,j)$,
then the strategy used by Algorithm} \ref{scoda}
{leads to an increase in modularity.
In the general case, we cannot control terms
$l_t(i,C(i))$ and $l_t(i,C(j))$,
but, in most cases, we expect the degree of attachment
of $i$ to $C(i)$ to be upper-bounded by some constant $\tau_1 < 1$ and
the degree of attachment of $i$ to $C(j)$ to be higher than some $\tau_2 > 0$.
Indeed, the fact that we observe an edge $(i,j)$ between node $i$ and and community $C(j)\neq C(i)$
is likely to indicate that the degree of attachment between $i$ and $C(j)$ is greater than what
we would have in the null model, and that the degree of attachment between $i$ and $C(i)$ is below maximum.
Moreover, since in real-world graphs the degree of most nodes is in $O(1)$ whereas $w$ is in $O(m)$,
we expect the term $(w_t(i) + 1)^2 / w$ to be smaller than a constant $\epsilon \ll 1$.
Then, the condition on $v_{\mathrm{max}}$ becomes:
}
\begin{equation*}
v_{\mathrm{max}} \leq \frac{1 - \epsilon}{\tau_1 - \tau_2}.
\end{equation*}
This justifies the design of the algorithm,
with the decision of joining one community or the other based on the community volumes.

\section{Experimental results}\label{experimental-results}
\label{section-experiments}

\subsection{Datasets}\label{datasets}

We use real-life graphs provided by the Stanford Social Network
Analysis Project (SNAP \citep{yang2015defining}) for the experimental
evaluation of {our new algorithm}. These datasets include ground-truth community
memberships that we use to measure the quality of the detection. {We
consider datasets of different natures.}
\textbf{Social networks}: The YouTube, LiveJournal, Orkut and
Friendster datasets correspond to social networks
\citep{backstrom2006group}\citep{mislove2007measurement} where nodes
represent users and edges connect users who have a friendship
relation. In all these networks, users can create groups that are used
as ground-truth communities in the dataset definitions.
\textbf{Co-purchasing network}: The Amazon dataset corresponds to a
product co-purchasing network \citep{leskovec2007dynamics}. The nodes
of the graph represent Amazon products and the edges correspond to
frequently co-purchased products. The ground-truth communities are
defined as the product categories.
\textbf{Co-citation network}: The DBLP dataset corresponds to a
scientific collaboration network \citep{backstrom2006group}. The nodes
of the graph represent the authors and the edges the co-authorship
relations. The scientific conferences are used as ground-truth
communities.

The size of these graphs ranges from approximately one million edges
to more than one billion edges. It enables us to test the ability of
{our algorithm} to scale to very large graphs. The characteristics of these
datasets can be found in Table \ref{table-snap-size-run-time}.


\subsection{Benchmark algorithms}\label{benchmark-algorithms}

For assessing the performance of {our streaming algorithm} we use a wide range of
state-of-the-art {but non-streaming} algorithms that are based on various approaches.
\textbf{SCD} (S) partitions the graph by maximizing the WCC, which is
a community quality metric based on triangle counting
\citep{prat2014high}.
\textbf{Louvain} (L) is based on the optimization of the well-known
modularity metric \citep{blondel2008fast}.
\textbf{Infomap} (I) splits the network into modules by compressing
the information flow generated by random walks
\citep{rosvall2008maps}.
\textbf{Walktrap} (W) uses random walks to estimate the similarity
between nodes, which is then used to cluster the network
\citep{pons2005computing}.
\textbf{OSLOM} (O) partitions the network by locally optimizing a
fitness function which measures the statistical significance of a
community \citep{lancichinetti2011finding}.
In the data tables, we use \textbf{STR} to refer to our streaming algorithm.

\subsection{Performance metrics and benchmark setup}\label{performance-metrics}

{We use two metrics for the performance evaluation of the selected
algorithms.} The first is the \emph{average F1-score}
\citep{yang2013overlapping}\citep{prat2014high} which corresponds to the harmonic
mean of precision and recall. The second metric is the \emph{Normalized Mutual Information}
(NMI), which is based on the mutual entropy between indicator functions
for the communities \citep{lancichinetti2009detecting}.

The experiments were performed on EC2 instances provided by Amazon Web
Services of type \texttt{m4.4xlarge} with 64 GB of RAM, 100 GB of disk
space, 16 virtual CPU with Intel Xeon Broadwell or Haswell and Ubuntu
Linux 14.04 LTS.

{Our algorithm} is implemented in C++ and the source code can be found on
GitHub\footnote{https://github.com/ahollocou/graph-streaming}. 
For the other algorithms, we used the C++ implementations provided by the authors,
that can be found on their respective websites. Finally, all the scoring
functions were implemented in C++. We used the implementation provided
by the authors of \citep{lancichinetti2009detecting} for the NMI and the
implementation provided by the authors of SCD \citep{prat2014high} for
the F1-Score.

\subsection{Benchmark results}\label{benchmark-results}

\subsubsection*{Execution time}\label{execution-time}
\addcontentsline{toc}{subsubsection}{Execution time}

We compare the execution times of the different algorithms on SNAP
graphs in Table \ref{table-snap-size-run-time}. The entries that are not
reported in the table corresponds to algorithms that returned execution
errors or algorithms with execution times exceeding \(6\) hours. In our
experiments, only SCD, except from {our algorithm}, was able to run on all
datasets. The fastest algorithms in our benchmarks are SCD and Louvain
and we observe that they run more than ten times slower than our
streaming algorithm. More precisely, {our streaming algorithm} runs in less than 50ms on the
Amazon and DBLP graphs, which contain millions of edges, and {less than 5}
minutes on the largest network, Friendster, that has more than one
billion edges. In comparison, it takes seconds for SCD and Louvain to
detect communities on the smallest graphs, and several hours to run on
Friendster. Table \ref{table-snap-size-run-time} {shows the execution times
of all the algorithms with respect to the number of edges in the
network}. We remark that there is more than one order of magnitude
between {our algorithm} and the other algorithms.

In order to compare the execution time of {our algorithm} with a minimal algorithm
that only reads the list of edges without doing any additional
operation, we measured the run time of the Unix command \texttt{cat} on
the largest dataset, Friendster. \texttt{cat} reads the edge file
sequentially and writes each line corresponding to an edge to standard
output. In our expermiments, the command \texttt{cat} takes 152 seconds
to read the list of edges of the Friendster dataset, whereas {our algorithm}
processes this network in {241} seconds. That is to say, reading the edge
stream is only twice faster than the execution of {our streaming algorithm}.

\begin{table}[h]
\begin{center}
\begin{tabular}{l | r r | r r r r r r}
& $|V|$ & $|E|$ & S & L & I & W & O & STR \\
\hline
Amazon      &    334,863 &       925,872   & 1.84 & 2.85 & 31.8 & 261 & 1038 & {\textbf{0.05}}\\
DBLP        &    317,080 &     1,049,866   & 1.48 & 5.52 & 27.6 & 1785 & 1717 & {\textbf{0.05}}\\
YouTube     &  1,134,890 &     2,987,624   & 9.96 & 11.5 & 150 & - & - & {\textbf{0.14}}\\
LiveJournal &  3,997,962 &    34,681,189   & 85.7 & 206 & - & - & - & {\textbf{2.50}}\\
Orkut       &  3,072,441 &   117,185,083   & 466 & 348 & - & - & - &  {\textbf{8.67}}\\
Friendster  & 65,608,366 & 1,806,067,135   & 13464 & - & - & - & -  & {\textbf{241}}\\
\end{tabular}
\end{center}
\caption{SNAP dataset sizes and execution times in seconds}
\label{table-snap-size-run-time}
\end{table}

\subsubsection*{Memory consumption}\label{memory-consumption}
\addcontentsline{toc}{subsubsection}{Memory consumption}

We measured the memory consumption of {streaming algorithm} and compared it to the
memory that is needed to store the list of the edges for each network,
which is a lower bound of the memory consumption of the other
algorithms. We use 64-bit integers to store the node indices. The memory
needed to represent the list of edges is 14,8 MB for the smallest
network, Amazon, and 28,9 GB for the largest one, Friendster. In
comparison, {our algorithm} consumes {8,1} MB on Amazon and only {1,6} GB on
Friendster.

\subsubsection*{Detection scores}\label{detection-scores}
\addcontentsline{toc}{subsubsection}{Detection scores}

{Table} \ref{table-snap-f1-score-nmi} shows the
Average F1-score and NMI of the algorithms on the SNAP datasets. Note
that the NMI on the Friendster dataset is not reported in the table
because the scoring program used for its computation
\citep{lancichinetti2009detecting} cannot handle the size of the output
on this dataset. While Louvain and OSLOM clearly outperform {our algorithm} on
Amazon and DBLP (at the expense of longer execution times), {our streaming algorithm} shows
similar performance as SCD on YouTube and much better performance than
SCD and Louvain on LiveJournal, Orkut and Friendster (the other
algorithms do not run these datasets). Thus {our algorithm} does not only run much
faster than the existing algorithms but the quality of the detected
communities is also better than that of the state-of-the-art algorithms
for very large graphs.

\begin{table}
\begin{center}
\begin{tabular}{l | r r r r r r | r r r r r r}
& \multicolumn{6}{c}{\textbf{F1-Score}}
& \multicolumn{6}{c}{\textbf{NMI}} \\
& S & L & I & W & O & STR
& S & L & I & W & O & STR \\
\hline
Ama.         & 0.39          & \textbf{0.47} & 0.30 & 0.39          & \textbf{0.47} & {0.38} 
             & 0.16          & 0.24          & 0.16 & \textbf{0.26} & 0.23          & {0.12} \\ 
DBLP         & 0.30          & 0.32          & 0.10 & 0.22          & \textbf{0.35} & {0.28} 
             & \textbf{0.15} & 0.14          & 0.01 & 0.10          & \textbf{0.15} & {0.10} \\ 
YT           & 0.23          & 0.11          & 0.02 & -             & -             & {\textbf{0.26}}
             & 0.10          & 0.04          & 0.00 & -             & -             & {\textbf{0.13}}\\
LiveJ.       & 0.19          & 0.08          & -    & -             & -             & {\textbf{0.28}}
             & 0.05          & 0.02          & -    & -             & -             & {\textbf{0.09}}\\
Orkut        & 0.22          & 0.19          & -    & -             & -             & {\textbf{0.44}}
             & 0.22          & 0.19          & -    & -             & -             & {\textbf{0.24}}\\
Friend.      & 0.10          & -             & -    & -             & -             & {\textbf{0.19}}
             & -             & -             & -    & -             & -             & -             \\
\end{tabular}
\end{center}
\caption{Average F1 Scores and NMI}
\label{table-snap-f1-score-nmi}
\end{table}


\section{Conclusion and future work}\label{conclusion-and-future-work}
\label{section-conclusion}

We introduced a new algorithm for the problem of graph clustering in the edge streaming setting.
In this setting, the input data is presented to the algorithm
as a sequence of edges that can be examined only once.
Our algorithm only stores three integers per node and
requires only one integer parameter $v_{\mathrm{max}}$.
It runs more than 10 times faster than state-of-the-art algorithms such as
Louvain and SCD and shows better detection scores on the largest
graphs. {Such an algorithm is extremely} useful in many applications where massive
graphs arise. {For instance,} the web graph contains around \(10^{10}\)
nodes which is much more than in the Friendster dataset.

We analyzed the adaptation of the popular modularity problem
to the streaming setting.
Theorem \ref{theorem-increase-condition} justifies the
nature of the condition on the volumes of the communities of nodes $i$ and $j$ for each new edge $(i,j)$,
which is the core of Algorithm \ref{scoda}.

{It would be interesting for future work to perform further experiments.
In particular the ability of the algorithm to handle evolving graphs could be
evaluated} on dynamic datasets \citep{panzarasa2009patterns}
{and compared to} existing approaches
\citep{gauvin2014detecting}\citep{epasto2015efficient}.
Note that, in the dynamic network settings,
modifications to the algorithm design could be made to handle events
such as edge deletions.

Finally, {our algorithm} only returns disjoint communities, whereas, in many real
graphs, overlaps between communities can be observed
\citep{lancichinetti2009detecting}. An important research direction
would consist in adapting {approach} to overlapping community detection and
compare it to existing approaches
\citep{xie2013overlapping}\citep{yang2013overlapping}.

\section*{Appendix A: Proof of Lemma 1}

Given a new edge $e_{t+1}=(i,j)$,
we have the following relation between quantities $\mathrm{Int}(C)$ and $\mathrm{Vol}(C)$
at times $t$ and $t+1$.
\begin{equation*}
\mathrm{Int}_{t+1}(C) = \mathrm{Int}_{t}(C) + 1_{i\in C}1_{j\in C}
\end{equation*}
and
\begin{equation*}
\mathrm{Vol}_{t+1}(C) = \mathrm{Vol}_t(C) + 1_{i\in C} + 1_{j\in C}.
\end{equation*}

This gives us the following equation for $(\mathrm{Vol}_{t+1}(C))^2$
\begin{equation*}
(\mathrm{Vol}_{t+1}(C))^2 = (\mathrm{Vol}_t(C))^2 +
\begin{cases}
0 & \text{if $C\neq C(i)$ and $C\neq C(j)$}\\
2 \mathrm{Vol}_t(C(i)) + 1 & \text{if $C=C(i)$}\\
2 \mathrm{Vol}_t(C(j)) + 1 & \text{if $C=C(j)$}\\
\end{cases}
\end{equation*}
in the case $C(i)\neq C(j)$, and
\begin{equation*}
(\mathrm{Vol}_{t+1}(C))^2 = (\mathrm{Vol}_t(C))^2 +
\begin{cases}
0 & \text{if $C\neq C(i)$ and $C\neq C(j)$}\\
4 \mathrm{Vol}_t(C(i)) + 4 & \text{if $C=C(i)=C(j)$}\\
\end{cases}
\end{equation*}
in the case $C(i)=C(j)$.

Finally, the definition of $Q_{t+1}$
\begin{equation*}
Q_{t+1} =
\sum_{C\in P_{t+1}} \left[ 2 \mathrm{Int}_{t+1}(C) - \frac{(\mathrm{Vol}_{t+1}(C))^2}{w}\right]
\end{equation*}
gives us the wanted result.

\section*{Appendix B: Proof of Lemma 2}

$Q_t$ is defined as a sum over all communities of partition $P_t$.
Only terms depending on $C(i)$ and $C(j)$ are modified by action (a).
Thus, we have:
\begin{equation*}
\begin{split}
\Delta Q_t &=
2 \left[
\mathrm{Int}_t(C(i)\setminus \{i\})
+ \mathrm{Int}_t(C(j)\cup \{i\})
- \mathrm{Int}_t(C(i))
- \mathrm{Int}_t(C(i)) \right]\\
&- \frac{
(\mathrm{Vol}_t(C(i)) - w_t(i))^2
+ (\mathrm{Vol}_t(C(j)) + w_t(i))^2
- (\mathrm{Vol}_t(C(i)))^2
- (\mathrm{Vol}_t(C(j)))^2
}{w}.
\end{split}
\end{equation*}

This leads to:
\begin{equation*}
\begin{split}
\Delta Q_t &=
2 \sum_{(i',j')\in S_t}
\left[ 1_{j'=i}(1_{i'\in C(j)} - 1_{i'\in C(i)})
+ 1_{i'=i}(1_{j'\in C(j)} - 1_{j'\in C(i)})\right]\\
&- 2 \frac{
w_t(i) \mathrm{Vol}_t(C(j))
- w_t(i) \mathrm{Vol}_t(C(i))
+ (w_t(i))^2
}{w}.
\end{split}
\end{equation*}

Using the definition of $\mathrm{Vol}_t$, we obtain the wanted expression for $\Delta Q_t$.

\section*{Appendix C: Proof of Theorem 1}

From Lemma \ref{lemma-Qt}, we obtain
\begin{equation*}
\begin{split}
\Delta Q_{t+1} &=
Q_t^{(a)} +
2 \left[ 1 - \frac{(\mathrm{Vol}_t(C(j)) + w_i(t)) + (\mathrm{Vol}_t(C(j)) + w_i(t)) + 2)}{w} \right] \\
&- Q_t^{(b)} -
2 \left[ 0 - \frac{(\mathrm{Vol}_t(C(i)) - w_i(t)) + (\mathrm{Vol}_t(C(j)) + w_i(t)) + 1)}{w} \right], \\
\end{split}
\end{equation*}
which gives us:
\begin{equation}
\label{equation-delta-Qt}
\Delta Q_{t+1} = \Delta Q_t +
2 \left[ 1 - \frac{\mathrm{Vol}_t(C(j)) - \mathrm{Vol}_t(C(i)) + 2 w_i(t) + 1}{w} \right]
\end{equation}

Then, Equation (\ref{equation-delta-Qt}) and Lemma \ref{lemma-Delta-Qt}
gives us the following expression for $\Delta Q_{t+1}$
\begin{equation*}
\begin{split}
\Delta Q_{t+1} =
2 \Bigg[ & 1
+ \left(l_t(i,C(j)) - \frac{1}{w}\right) Vol_t(C(j))
- \left(l_t(i,C(i)) - \frac{1}{w}\right) Vol_t(C(i))\\
&-
\frac{(w_t(i) + 1)^2}{w} \Bigg].
\end{split}
\end{equation*}

Thus, $\Delta Q_{t+1} \geq 0$ is equivalent to
\begin{equation}
\label{equation-inequality-ut}
\left(l_t(i,C(i)) - \frac{1}{w}\right) Vol_t(C(i))\\
- \left(l_t(i,C(j)) - \frac{1}{w}\right) Vol_t(C(j))
\leq 1 -
\frac{(w_t(i) + 1)^2}{w}.
\end{equation}

We use $u_t(i,j)$ to denote the left-hand side of this inequality.
If $Vol_t(C(i)) \leq Vol_t(C(j))$, then we have
\begin{equation*}
u_t(i,j) \leq [l_t(i, C(i)) - l_t(i,C(j))] Vol_t(C(j))
\end{equation*}

Thus, the following inequality
\begin{equation*}
[l_t(i, C(i)) - l_t(i,C(j))] Vol_t(C(j)) \leq
1 - \frac{(w_t(i) + 1)^2}{w}
\end{equation*}
implies inequality (\ref{equation-inequality-ut}), which proves the theorem.

\end{document}